\pdfoutput=1
\documentclass[11pt]{article}

\usepackage[preprint]{acl}

\usepackage{times}
\usepackage{latexsym}
\usepackage[T1]{fontenc}
\usepackage[utf8]{inputenc}
\usepackage{microtype}
\usepackage{inconsolata}
\usepackage{graphicx}
\graphicspath{{figures/}} %
\usepackage{flafter} %
\usepackage{stfloats} %
\usepackage{capt-of} %
\usepackage{booktabs}
\usepackage{amssymb}
\usepackage{amsmath}
\usepackage[most]{tcolorbox}

\usepackage{tikz}
\usetikzlibrary{arrows.meta, positioning, calc, backgrounds, fit}
\definecolor{okBlue}{HTML}{0072B2}
\definecolor{okOrange}{HTML}{E69F00}

\usepackage{listings}
\lstdefinestyle{yamlpaper}{
  basicstyle=\ttfamily\scriptsize,
  backgroundcolor=\color{gray!4},
  frame=single,
  rulecolor=\color{gray!40},
  framesep=5pt,
  xleftmargin=6pt,
  breaklines=true,
  columns=fullflexible,
  keepspaces=true,
  showstringspaces=false,
  morecomment=[l]{\#},
  commentstyle=\color{gray!55!black}\itshape,
  captionpos=b,
}

\title{Agentic AutoRAG: RAG Pipeline Optimization \\ through Reasoning-Driven Agents}

\author{
  \textbf{Lasse B. Strand}\thanks{Corresponding author: \texttt{lstrand.research@gmail.com}},
  \textbf{Robert Jakob},
  \textbf{Kevin O'Sullivan},
  \textbf{Markus Kreft}
\\
  ETH Zurich
}

\begin{document}
\maketitle
\begingroup\renewcommand{\thefootnote}{}\footnotetext{Accepted at the Second Workshop for REsearch on Agent Language Models (REALM) at EMNLP 2026 and at the Machine Learning for Systems Workshop at NeurIPS 2026.}\endgroup

\begin{abstract}
Retrieval-augmented generation (RAG) is a widely used approach for grounding large language models (LLMs) in external knowledge. However, configuring a pipeline is an expensive hyperparameter optimization problem over many interacting choices, from chunking and embedding model to reranking and generation. Existing optimizers, from greedy search to Bayesian optimization, reduce each trial to an aggregate score and search without modeling why a configuration performed as it did, even though the retrieved chunks already provide evidence about whether each failure occurred during retrieval or after it. We introduce Agentic AutoRAG, an LLM-agent optimizer for multi-objective RAG hyperparameter optimization with retrieval-versus-generation failure attribution. It proposes configurations scored on a frozen exam from the corpus: after each trial a Diagnoser attributes each failed question to retrieval or generation, and a Proposer, grounded in a knowledge base of model rankings and pricing, selects the next configuration, weighing accuracy against cost to trace a Pareto frontier. On three multi-hop QA benchmarks it reaches higher LLM-judge accuracy than every baseline we compare, and within its first 10 trials it matches or beats the statistical baselines' full 30-trial judge accuracy. In its cost-aware mode on a real-world healthcare corpus it reaches a median exam accuracy of 77\%, above the strongest baseline's 71.5\%, at about 58\% of that baseline's cost per query, and it matches that 71.5\% at about 22\% of the cost.\footnote{Code: \url{https://github.com/Agentic-Systems-Lab/Agentic-AutoRAG}}
\end{abstract}

\section{Introduction}
\label{sec:intro}

Retrieval-augmented generation (RAG) has become a standard way to ground large language models (LLMs) in knowledge beyond their training data \citep{lewis2020rag, gao2023ragsurvey}. Configuring a RAG pipeline means choosing across many interacting dimensions: chunking, embedding model, index type, retrieval depth, reranker, query expansion, passage compression, and generator. These choices are coupled and should not be tuned in isolation \citep{wang2024bestpractices}. The best retrieval depth, for instance, depends on chunk size: small chunks each hold less text, so they need a deeper retrieval to surface the same evidence that a few large chunks already contain. The best configuration also shifts from corpus to corpus: a pipeline tuned for one can underperform on another, so the search must be repeated for each new one \citep{kartal2025ragsmith, orbach2025comparing}. Scoring a configuration means running the whole pipeline over a test set, so the search must be sample-efficient.

Recent work applies general-purpose search algorithms to this problem: AutoRAG's greedy node-wise selection \citep{kim2024autorag}, multi-armed bandits \citep{fu2024autoraghp}, Bayesian optimization \citep{conway2025syftr, barker2025faster}, and genetic search \citep{kartal2025ragsmith}. All of these optimizers reduce each trial to an aggregate score and search the space without modeling why a configuration performed as it did. When a configuration scores low, they cannot tell whether retrieval and reranking failed to surface the relevant passages, or the generator answered incorrectly despite having the right context. However, any RAG pipeline already records which chunks it retrieved. By checking whether the passages that contain the answer (the gold spans) appear among them, we can attribute each failure to retrieval or generation. We are not aware of prior RAG optimizers that use this per-question failure evidence to drive their configuration proposals.

We introduce Agentic AutoRAG, a reasoning-driven optimizer for RAG pipelines. Like recent LLM-agent optimizers for general hyperparameter tuning \citep{liu2025agenthpo, trirat2025automlagent}, it uses a language model to propose configurations, but specializes this approach to RAG and cost-aware search. It contributes three ideas:

\begin{enumerate}
\item A \textbf{diagnose-then-propose loop.} After each trial, a Diagnoser reads the per-question failure evidence, attributes each failure to retrieval or generation, and compresses it into a short diagnosis. A Proposer then reads that diagnosis and selects the next configuration.

\item A \textbf{knowledge-base-grounded, cost-aware agent.} The Proposer prices and ranks candidate configurations against public model leaderboards and per-token pricing before spending a trial on them, so it navigates the accuracy-cost trade-off without first learning which models are strong or cheap from trial outcomes.

\item A \textbf{corpus-derived exam.} Adapting established corpus-grounded exam generation \citep{guinet2024examgen, shashidhar2025yourbench}, we draft open-ended, multi-hop questions with verbatim gold-answer spans, filter them through a sufficiency oracle, and keep only questions that separate configurations spanning the search space. The exam is frozen once and reused as the optimization target across every trial.
\end{enumerate}

We compare the optimizer against random search and a multi-objective Bayesian optimizer (MO-TPE). On three multi-hop QA benchmarks it reaches higher LLM-judge accuracy than every baseline we compare, and it is sample-efficient: within its first 10 trials it already matches or beats the statistical baselines' full 30-trial judge accuracy. In its cost-aware mode on a healthcare corpus it reaches a median exam accuracy of 77\%, above the strongest baseline's 71.5\%, at about 58\% of that baseline's cost per query, and it matches that 71.5\% at about 22\% of the cost.

\section{Related Work}
\label{sec:related}

\subsection{RAG pipeline optimization}
\label{sec:related:rag-opt}

A RAG pipeline exposes many discrete choices, making its configuration a hyperparameter optimization (HPO) problem with an expensive black-box objective.

AutoRAG \citep{kim2024autorag} decomposes the pipeline into modular nodes and greedily selects the best module at each node in pipeline order. AutoRAG-HP \citep{fu2024autoraghp} casts RAG tuning as a multi-armed bandit problem and tunes three hyperparameters with a hierarchical UCB1 variant. syftr \citep{conway2025syftr} applies multi-objective Bayesian optimization (MO-TPE) to a large space of RAG pipelines, from standard to agentic, with a Pareto-Pruner that halts candidates unlikely to extend the frontier and a transfer step that seeds a new search with configurations found on other datasets. \citet{barker2025faster} add more objectives, tuning cost, latency, safety, and alignment jointly with multi-objective Bayesian optimization. RAGSmith \citep{kartal2025ragsmith} searches RAG compositions with a genetic algorithm. \citet{orbach2025comparing} compare TPE, random search, and greedy variants and find that about 10 trials reach the accuracy of an exhaustive grid, though over a space of only 162 configurations, far smaller than the one we search.

All of these methods reduce a trial to its score and do not model why a configuration failed, which is the per-question evidence our optimizer diagnoses after each trial to guide its next proposal.

\subsection{LLM agents as optimizers}
\label{sec:related:llm-opt}

A parallel line of work uses an LLM as the optimizer itself, though not yet applied to RAG pipeline configuration. OPRO \citep{yang2024opro} gives the model a meta-prompt that lists the solutions tried so far with their scores and asks it to propose a better one, shown on tasks such as linear regression and prompt optimization.

In HPO specifically, \citet{liu2025agenthpo} introduce AgentHPO, a two-agent framework where a Creator proposes hyperparameters and an Executor runs experiments across 12 ML tasks. \citet{trirat2025automlagent} apply LLM agents to a broader range of AutoML tasks.

These optimizers are general-purpose and single-objective, and they carry no prior about which models are strong or cheap. We specialize the approach to RAG pipeline configuration, and we ground the agent in a knowledge base of public model rankings and per-token pricing so it can weigh accuracy against cost and rule out configurations before running them.

A separate line of work builds agentic RAG systems, where an agent decides what to retrieve while answering each question \citep{asai2024selfrag, yu2024autorag, wu2025composerag, leng2025decexrag, singh2025agenticrag}. Our agent instead tunes the pipeline configuration rather than answering questions itself.

\subsection{Automated RAG evaluation}
\label{sec:related:eval}

Building an evaluation set from the target corpus is an established way to assess retrieval-augmented systems without hand annotation \citep{guinet2024examgen, saadfalcon2024ares, shashidhar2025yourbench, filice2025datamorgana}. Closest to us, \citet{guinet2024examgen} prompt an LLM to write corpus-grounded multiple-choice questions and use item response theory to discard questions that do not separate systems. Our exam adapts this recipe for configuration search: the questions are open-ended with verbatim gold spans, and we keep only those that separate configurations spanning our search space.

\section{Method}
\label{sec:method}

\begin{figure*}[t]
\centering
\begin{tikzpicture}[
  x=1mm, y=1mm, font=\footnotesize,
  box/.style={rounded corners=2pt, align=center, text width=19mm,
    minimum height=16mm, inner sep=2.5pt, line width=0.5pt,
    draw=black!60, fill=white, text=black!88,
    execute at begin node={\hyphenpenalty=10000\relax}},
  flow/.style={-{Stealth[length=2mm]}, line width=0.6pt, draw=black!62,
    rounded corners=1.6pt},
  grp/.style={rounded corners=3pt, line width=0.7pt},
  gtitle/.style={font=\small\bfseries},
]
  \node[box, text width=15mm, minimum height=10mm] (corpus) at (12,6)   {Corpus};
  \node[box, text width=15mm, minimum height=10mm] (search) at (12,-6)  {Search space};
  \node[box, text width=26mm, minimum height=12mm] (kb)     at (12,-51)
    {Knowledge Base (model specs \& benchmark scores)};

  \node[box] (bn)    at (45.25,0)   {Build neighborhoods};
  \node[box] (comp)  at (69.25,0)   {Composer Agent};
  \node[box] (val)   at (93.25,0)   {Validate (filters + oracle)};
  \node[box] (probe) at (117.25,0)  {Probe-based discrimination};
  \node[box, text width=16mm, minimum height=14mm] (fx) at (147,0) {Frozen exam};

  \node[box] (pini) at (45.25,-43)  {Proposer Agent suggests initial config};
  \node[box] (eval) at (69.25,-43)  {Evaluate RAG config on exam};
  \node[box] (diag) at (93.25,-43)  {Diagnoser Agent};
  \node[box] (prop) at (117.25,-43) {Proposer Agent};
  \node[box, text width=19mm, minimum height=20mm] (out) at (147,-43)
    {Recommended config + Pareto frontier + Reasoning trace};

  \begin{scope}[on background layer]
    \node[grp, draw=black!40, fill=black!4, densely dashed,
      fit=(corpus)(search), inner xsep=3mm, inner ysep=4mm] (up){};
    \node[grp, draw=okBlue!55, fill=okBlue!6,
      fit=(bn)(probe), inner xsep=5mm, inner ysep=5mm] (examg){};
    \node[grp, draw=okOrange!70!black, fill=okOrange!8,
      fit=(pini)(prop), inner xsep=5mm, inner ysep=6mm] (loopg){};
  \end{scope}
  \node[gtitle, text=black!55, anchor=south west] at (up.north west)  {User Provided};
  \node[gtitle, text=okBlue!45!black, anchor=south] at (examg.north)  {Exam Generation};
  \node[gtitle, text=okOrange!60!black, anchor=south] at (loopg.north) {Optimization Loop};

  \draw[flow] (up.east) -- (examg.west);
  \draw[flow] (bn) -- (comp);
  \draw[flow] (comp) -- (val);
  \draw[flow] (val) -- (probe);
  \draw[flow] (examg.east) -- (fx.west);

  \draw[flow] (pini) -- (eval);
  \draw[flow] (eval) -- (diag);
  \draw[flow] (diag) -- (prop);
  \draw[flow] (loopg.east) -- (out.west);

  \draw[flow] (fx.south) -- (147,-20) -- (18,-20) -- (18,-35)
              -- ([yshift=8mm]loopg.west);                    %
  \draw[flow] (up.south) -- (12,-43) -- (loopg.west);         %
  \draw[flow] (kb.east) -- ([yshift=-8mm]loopg.west);         %

  \draw[flow] (prop.south) -- (117.25,-54) -- (69.25,-54) -- (eval.south);
  \node[fill=okOrange!8, inner sep=1.5pt, font=\scriptsize\itshape, text=black!70]
    at (93.25,-54) {Until max trials reached};
\end{tikzpicture}
\caption{Architecture of Agentic AutoRAG. The exam generation pipeline builds a frozen exam from the corpus once before optimization: TF-IDF neighborhoods seed the Composer, which drafts open-ended questions with verbatim gold spans, validation filters and a sufficiency oracle enforce answerability and multi-hop dependence, and probe-based discrimination admits only questions that differentiate across the search space. The optimization loop alternates between a Diagnoser, which compresses each trial's per-question failure evidence, and a Proposer, which selects the next configuration grounded in a knowledge base of public model rankings and pricing. The optimizer outputs a Pareto frontier over accuracy and cost, a recommended configuration, and a per-trial reasoning trace.}
\label{fig:architecture}
\end{figure*}
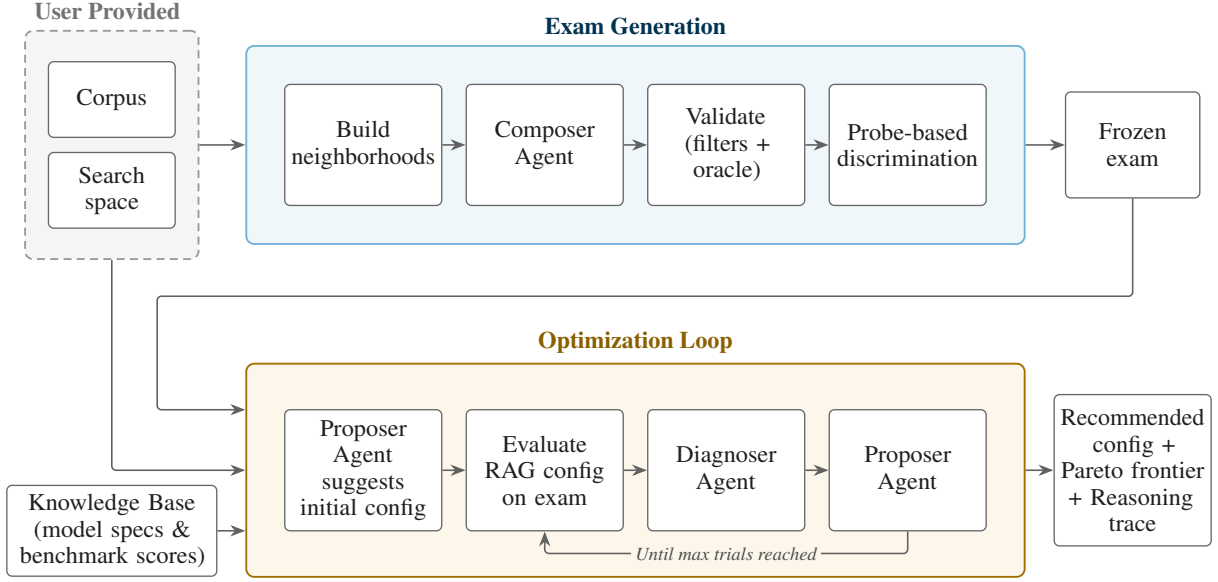

Agentic AutoRAG is a reasoning-driven optimizer for RAG pipelines. Each trial runs a candidate configuration on a frozen exam of question-answer pairs, and a two-stage reasoning loop interprets the result. A Diagnoser interprets per-question failure evidence from the trial that just ran, and a Proposer selects the next configuration based on the diagnosis. The optimizer searches over answer accuracy and cost per query and produces a Pareto frontier of non-dominated configurations. Figure~\ref{fig:architecture} gives an overview.

\subsection{Problem Formulation}
\label{sec:method:formulation}

A RAG pipeline $P_{\mathbf{c}}$ maps a question $q$ to an answer $a$ under a configuration $\mathbf{c}$ drawn from a user-defined search space $\mathcal{S}$. The space is a product of discrete and continuous dimensions covering chunking, embedding, index type, retrieval, reranking, query expansion, passage compression, and generation.

We optimize two objectives: answer accuracy $f(\mathbf{c}) \in [0,1]$, the fraction of exam questions answered correctly, and mean API cost per query $g(\mathbf{c})$, in USD. A configuration $\mathbf{c}^*$ Pareto-dominates $\mathbf{c}$ when $f(\mathbf{c}^*) \geq f(\mathbf{c})$ and $g(\mathbf{c}^*) \leq g(\mathbf{c})$ with at least one strict inequality. The optimizer runs in one of two modes. In its default cost-aware mode it optimizes both objectives and returns the non-dominated set of configurations found across trials. In score-only mode it optimizes accuracy alone and hides the cost and Pareto signals from the agents.

\subsection{Exam Generation}
\label{sec:method:examiner}

Our framework builds an open-ended exam of question-answer pairs from the target corpus once before optimization and reuses it across every trial, so scores stay comparable from trial to trial. Exam generation is an optional component: a user can supply a question-answer set directly, in which case the system skips generation and scores each trial on the supplied exam. When it does generate the exam, an LLM agent we call the Composer drafts the questions in three stages: it reads neighborhoods of related chunks and drafts questions from them, validation filters the drafts, and probe-based selection keeps the questions that separate one configuration from another. The Composer over-generates by a configurable factor, so the candidate pool is larger than the target exam, and validation and selection then reduce it to the requested number of questions.

\paragraph{Neighborhoods.}
For each generation call the Composer reads a neighborhood of related chunks and drafts as many questions as it judges the material supports, tagging each with a question type from a fixed taxonomy of seven types (Appendix~\ref{sec:appendix:taxonomy}, with an example of each in~\ref{sec:appendix:examples}). A neighborhood pairs an anchor chunk with its same-document siblings and a set of cross-document chunks. The cross-document chunks are ranked by TF-IDF cosine similarity to the anchor, a lexical score that favors chunks sharing rare, distinctive terms with the anchor rather than common words.

\paragraph{Validation.}
Each candidate passes a set of deterministic filters: a structural check that rejects a question citing one span twice rather than two distinct ones, a self-containment filter that rejects questions referring to their source by name (``the document'', ``the paper'', ``Chunk 1''), a non-empty-span check, and an arithmetic verification for numeric questions. Multi-hop questions then pass a sufficiency oracle. When shown all gold spans at once, the oracle must return the smallest subset of spans that supports the answer, a verbatim quote from each, and the answer itself. A question is rejected when that subset is smaller than its hop count, since it would then decompose into fewer hops. Testing the spans one at a time would misjudge sufficiency, because a single span shown alongside the question often lets the oracle infer the answer from the question itself and rate that span sufficient on its own.

\paragraph{Probe-based selection.}
A question is useful to the optimizer only when some of the configurations it will evaluate solve the question and others do not. We measure this with four probe configurations that span the search-space extremes along several axes (LLM, embedding model, reranker, chunk size, and retrieved top-$k$), and we evaluate every candidate on all four. We discard the questions that every probe answers, since they no longer separate one configuration from another, favor the questions that only one or a few probes solve, and keep a small share that no probe solves, since a configuration beyond the probe envelope may still answer them. This weighting governs which questions enter the exam, not how answers are later scored.

\subsection{Optimization Loop}
\label{sec:method:loop}

On each trial the system builds or loads a cached index for the proposed configuration, runs the RAG pipeline on every exam question, scores the answers, and records which gold spans appear in the retrieved chunks. Recording the retrieved spans is what lets the loop attribute a failure to retrieval, when the gold spans were missed, or to generation, when they were retrieved but the answer was still wrong. The trial's outcome passes to two reasoning agents, a Diagnoser and a Proposer.

\paragraph{Diagnoser.}
The Diagnoser compresses the trial's per-question evidence into a compact diagnosis, which keeps the Proposer's context window focused on what matters. It reads a pre-computed summary of the trial that just ran: overall accuracy and accuracy on the questions where retrieval succeeded, how the failures divide between retrieval and generation, and a sample of failed questions chosen to span different failure types and question types, including questions this configuration newly got wrong, rather than only the lowest-scoring ones. It also sees a compact record of how accuracy moved across earlier trials, but not their configurations, their cost, the Pareto frontier, or its own earlier diagnoses. Withholding these keeps each diagnosis grounded in the current trial and leaves synthesis across trials to the Proposer. The Diagnoser emits a short narrative, findings tied to the trial's numbers, notable changes from earlier trials, and a handful of illustrative questions for the Proposer. It does not prescribe configuration changes.

\paragraph{Proposer.}
The Proposer chooses the next configuration. It reads the diagnosis, the current configuration, the search space with its legal ranges, and a knowledge base of model rankings and pricing. A short plan, carried forward from one trial to the next, serves as the loop's memory. Every configuration evaluated so far stays in view, so it never repeats one. It also knows which generators, embedding models, and rerankers remain untried, and how much of the trial budget is left. In cost-aware mode it reads the current accuracy-cost frontier directly, with each non-dominated member's accuracy, cost, and configuration, and decides where along the cost axis to probe next. It emits the next configuration together with a rationale.

The knowledge base lets the Proposer rule out configurations that public benchmarks predict will underperform before a trial is spent on them. This is what makes the loop tractable in a large search space, where naive enumeration is infeasible and Bayesian baselines spend trials rediscovering the same priors. More importantly, it is what enables cost-aware search, as its per-token pricing lets the Proposer weigh accuracy against cost before each trial and move straight to cheap-but-capable generators. Appendix~\ref{sec:appendix:kb} details its sources and the per-model annotations the Proposer reasons over.

The optimizer runs for a user-specified number of trials. In cost-aware mode it returns the Pareto frontier together with an agent-recommended configuration from it, and in score-only mode the single best-scoring configuration. Either way it also returns the per-trial reasoning trace.

\section{Experimental Setup}
\label{sec:setup}
We evaluate Agentic AutoRAG in two experiments. The first, the Accuracy experiment, maximizes answer accuracy on three multi-hop question-answering benchmarks and compares our optimizer against random search, MO-TPE with and without a warm start, and a knowledge-and-diagnosis ablation of our own agent. To compare the optimizer independently of the exam-generation component of our framework, this experiment optimizes against a fixed exam built from each benchmark's own labeled questions rather than the corpus-derived exam of Section~\ref{sec:method:examiner}. The second, the Pareto experiment, turns on the optimizer's cost-aware mode and runs the full pipeline end to end, generating an exam from a corpus of real-world healthcare documents and then tracing the accuracy-cost Pareto frontier it finds. Every method runs under a fixed budget of 30 trials, and we repeat each search ten times with independent seeds.

\paragraph{Datasets.}
The Accuracy experiment uses three multi-hop question-answering benchmarks. HotpotQA \citep{yang2018hotpotqa} in the distractor setting asks questions that bridge or compare two Wikipedia paragraphs. MuSiQue-Answerable \citep{trivedi2022musique} composes two to four single-hop questions into one, so many items need a longer reasoning chain. MultiHop-RAG \citep{tang2024multihoprag} draws comparison, inference, and temporal questions from a news corpus and includes unanswerable questions whose correct response is to abstain. The Pareto experiment uses the healthcare subset of UniDoc-Bench \citep{peng2025unidocbench}, 230 PDFs and 20 page images that are mostly biomedical research articles and medical studies. These documents are multi-page, with tables, figures, and mixed layouts, and we parse them with Docling \citep{auer2024doclingtechnicalreport} using table-structure extraction and OCR. We use the documents as a corpus, generate the exam from them with our own pipeline, and do not use UniDoc-Bench's labeled question-answer pairs. Table~\ref{tab:corpora} reports corpus sizes.

\begin{table}[t]
\centering
\small
\begin{tabular}{@{}lrrr@{}}
\toprule
Corpus & Docs. & Words & Words/doc. \\
\midrule
HotpotQA & 19{,}276 & 1.81M & 94 \\
MuSiQue & 17{,}629 & 1.32M & 75 \\
MultiHop-RAG & 609 & 1.07M & 1{,}760 \\
\midrule
UniDoc & 250 & 1.30M & 5{,}214 \\
\bottomrule
\end{tabular}
\caption{Corpus size for each benchmark.}
\label{tab:corpora}
\end{table}

\paragraph{Search space.}
All methods search the same space of pipeline configurations. It spans 2 chunking strategies over 4 chunk sizes and 4 overlaps, 13 embedding models, 2 index types with continuous retrieval depth and fusion weight, 4 query-expansion strategies, 5 reranker choices with continuous rerank depth, and 35 generator LLMs with a per-trial reasoning toggle. Appendix~\ref{sec:appendix:searchspace} lists every dimension with its range.

\begin{figure*}[t]
\centering
\includegraphics[width=\textwidth]{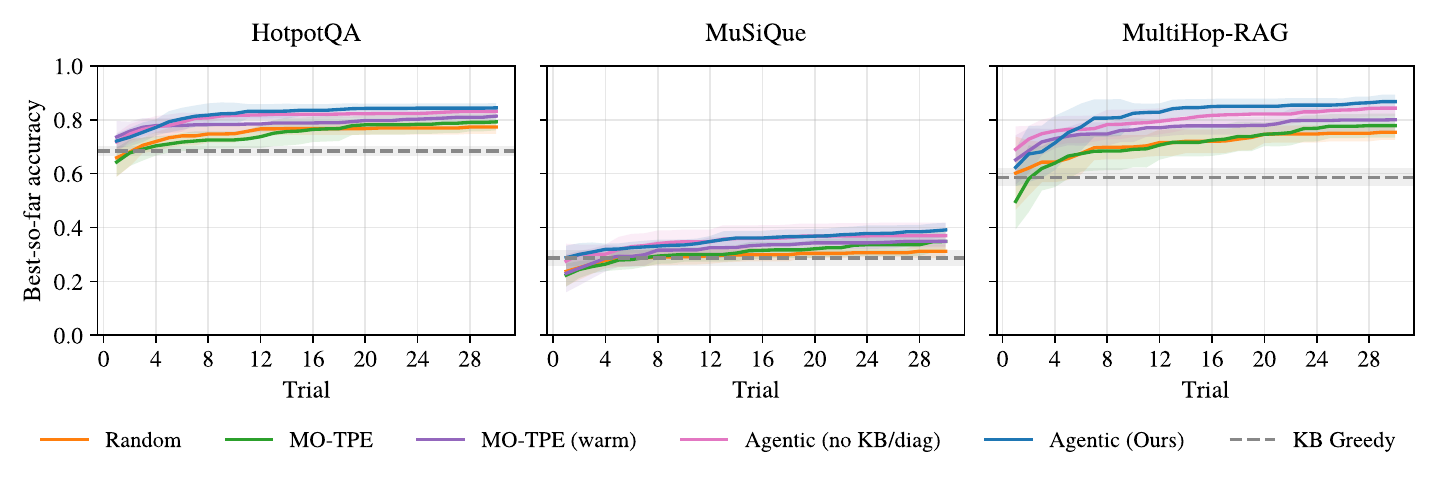}
\caption{Best-so-far validation-exam accuracy over the 30-trial search, one panel per dataset. Each line is a method's running-best accuracy averaged over the 10 search seeds, and the shaded band spans $\pm 1$ standard deviation across seeds. KB Greedy, the no-search reference, is the dashed horizontal line.}
\label{fig:best-so-far}
\end{figure*}

\paragraph{Optimizers.}
Every method sees the same search space and is scored on the same fixed exam at each trial. Random search samples configurations uniformly and resamples when a draw violates a dependency constraint. MO-TPE is the multi-objective Tree-structured Parzen Estimator from Optuna \citep{akiba2019optuna} that recent RAG-tuning work converges on \citep{conway2025syftr, orbach2025comparing}. We follow syftr \citep{conway2025syftr} in using it as the search engine and replicate it in our own harness with a multivariate TPE sampler. In the Accuracy experiment it optimizes answer accuracy alone, and in the Pareto experiment it optimizes accuracy and cost jointly. We do not run the full syftr system: we restrict its space to dense and hybrid pipelines, leaving out the agentic-RAG variants it also searches, disable its Pareto-frontier pruning, and run 30 trials per search rather than the several hundred syftr uses. Reduced to its search engine, MO-TPE differs from our agent only in how it chooses each configuration, so the comparison isolates the search strategy. MO-TPE (warm) additionally seeds the study with the 30 completed trials of the paired random-search run on the same corpus as a free, uncounted prior, so it draws on information from 60 evaluations while every other method sees 30. We report it at the same 30 counted trials as a strong reference.\footnote{This is a same-corpus warm start. It is not the cross-dataset transfer seeding of syftr, which starts a new search from Pareto-optimal flows found on different datasets.} We also run an ablation of our own agent, Agentic (no KB/diag), that removes both the knowledge base and the per-question diagnosis and leaves an LLM optimizer that sees only the score trajectory and its own past configurations. The gap between it and the full agent therefore measures what the diagnosis loop and knowledge base add on top of the language model. As a non-search reference, KB Greedy takes each dimension's top-rated option directly from the knowledge base without running any trial. Our optimizer runs in score-only mode in the Accuracy experiment and in cost-aware mode in the Pareto experiment.

\paragraph{Evaluation and metrics.}
In the Accuracy experiment, every method is scored at each trial on a fixed 100-question validation exam built from the benchmark's labeled questions, and each returned configuration is then re-evaluated on a separate 300-question held-out gold test set that is disjoint from the search questions and unseen during search. Both question sets are identical for every method and every seed, and the seeds vary only the optimizer's own randomness, not the questions. Because each seed still returns its own configuration, this gives ten held-out evaluations per method, and we report the mean and standard deviation across them. For the agentic optimizer we also take its best configuration within the first 10 and 20 trials as two additional returned configurations, reported as Agentic@10 and Agentic@20. The validation exam and the held-out set are different questions, so per-trial validation scores and held-out scores are not directly comparable. On the held-out set we report exact match, token-level F1, and LLM-judge accuracy against the gold answers. We also report each method's search cost and embedding compute in Appendix~\ref{sec:appendix:cost}. In the Pareto experiment we report the Pareto frontier over answer accuracy and mean per-query API cost, scored on the corpus-derived exam, which is generated once before any search and identical for every method and every seed. For each query we sum the price of its generation and query-expansion calls from LiteLLM's model catalog and average over queries. Embedding and reranking models run locally at no API cost, so this axis is the generator-side cost, which also depends on retrieval because the retrieved passages become the generator's input tokens.

\begin{table*}[t]
\centering
\footnotesize
\setlength{\tabcolsep}{1.5pt}
\begin{tabular}{@{}l ccc ccc ccc@{}}
\toprule
& \multicolumn{3}{c}{HotpotQA} & \multicolumn{3}{c}{MuSiQue} & \multicolumn{3}{c}{MultiHop-RAG} \\
\cmidrule(lr){2-4} \cmidrule(lr){5-7} \cmidrule(lr){8-10}
Method & EM & F1 & Judge & EM & F1 & Judge & EM & F1 & Judge \\
\midrule
Random & $.506_{\pm .038}$ & $.664_{\pm .041}$ & $.774_{\pm .046}$ & $.165_{\pm .029}$ & $.215_{\pm .035}$ & $.270_{\pm .036}$ & $.633_{\pm .065}$ & $.651_{\pm .063}$ & $.706_{\pm .048}$ \\
MO-TPE & $.515_{\pm .057}$ & $.673_{\pm .056}$ & $.784_{\pm .056}$ & $.193_{\pm .024}$ & $.254_{\pm .032}$ & $.301_{\pm .032}$ & $.664_{\pm .045}$ & $.677_{\pm .042}$ & $.724_{\pm .028}$ \\
MO-TPE (warm) & $.521_{\pm .044}$ & $.677_{\pm .055}$ & $.786_{\pm .057}$ & $.183_{\pm .030}$ & $.243_{\pm .032}$ & $.301_{\pm .039}$ & $.666_{\pm .051}$ & $.682_{\pm .047}$ & $.731_{\pm .045}$ \\
Agentic (no KB/diag) & $.550_{\pm .020}$ & $.711_{\pm .024}$ & $.819_{\pm .025}$ & $.180_{\pm .022}$ & $.230_{\pm .028}$ & $.298_{\pm .024}$ & $\mathbf{.729}_{\pm .037}$ & $\mathbf{.748}_{\pm .035}$ & $.780_{\pm .021}$ \\
Agentic@10 & $.558_{\pm .015}$ & $.715_{\pm .017}$ & $.817_{\pm .014}$ & $.177_{\pm .023}$ & $.231_{\pm .028}$ & $.303_{\pm .030}$ & $.679_{\pm .056}$ & $.697_{\pm .055}$ & $.773_{\pm .021}$ \\
Agentic@20 & $\mathbf{.568}_{\pm .012}$ & $\mathbf{.727}_{\pm .009}$ & $\mathbf{.835}_{\pm .013}$ & $\mathbf{.198}_{\pm .027}$ & $.261_{\pm .028}$ & $.326_{\pm .030}$ & $.704_{\pm .060}$ & $.723_{\pm .061}$ & $.783_{\pm .026}$ \\
Agentic (Ours) & $.566_{\pm .015}$ & $.723_{\pm .014}$ & $.825_{\pm .015}$ & $.196_{\pm .031}$ & $\mathbf{.262}_{\pm .028}$ & $\mathbf{.330}_{\pm .025}$ & $\mathbf{.729}_{\pm .054}$ & $\mathbf{.748}_{\pm .055}$ & $\mathbf{.790}_{\pm .029}$ \\
\midrule
KB Greedy & $.480_{\pm .016}$ & $.611_{\pm .016}$ & $.702_{\pm .019}$ & $.150_{\pm .008}$ & $.187_{\pm .008}$ & $.254_{\pm .014}$ & $.507_{\pm .011}$ & $.518_{\pm .011}$ & $.544_{\pm .014}$ \\
\bottomrule
\end{tabular}
\caption{Held-out accuracy on each benchmark's gold test slice, a 300-question stratified sample disjoint from the validation exam and unseen during search. Exact match, token-level F1, and LLM-judge accuracy are reported as mean$_{\pm\text{SD}}$ over the 10 search seeds. Rows above the rule are search methods run under a 30-trial budget. Agentic@10 and @20 are the agentic optimizer's best configuration within its first 10 and 20 trials. KB Greedy, below the rule, is a no-search reference that takes each dimension's top option from the knowledge base. Best value per column among the search methods is in bold, with ties bold on both rows.}
\label{tab:holdout}
\end{table*}

\section{Results}
\label{sec:results}

\subsection{Accuracy experiment}
\label{sec:results:accuracy}

\paragraph{Best-so-far accuracy.}
On all three datasets (Figure~\ref{fig:best-so-far}) the agentic optimizer's best-so-far accuracy climbs fastest, reaching most of its final accuracy within the first few trials and staying at the top of every panel through the rest of the budget. Random search and cold MO-TPE climb more slowly and settle lower, and warm-started MO-TPE sits between them and the agent. The no-KB/diag ablation tracks closest to the full agent, so most of this early efficiency comes from using a language model to propose configurations. The added lift from the knowledge base and per-question diagnosis is small in the trace and shows up more clearly in the held-out numbers.

\paragraph{Held-out accuracy.}
On the held-out gold test set (Table~\ref{tab:holdout}) the agentic optimizer achieves higher LLM-judge accuracy than every baseline across all three benchmarks, at 0.825 on HotpotQA, 0.330 on MuSiQue, and 0.790 on MultiHop-RAG, above random search and with the two MO-TPE variants in between. Exact match and F1 tell the same story on HotpotQA and MuSiQue, where an agentic run holds the top score in every column. On MultiHop-RAG the no-KB/diag ablation matches the full agent's exact match and F1 (0.729 and 0.748) and trails it only slightly on the judge, so the knowledge base and per-question diagnosis add little there over a plain LLM optimizer. Accuracy is low for every method on MuSiQue, whose up-to-four-hop questions need multi-step retrieval that conditions each hop on the previous hop's results, a pipeline shape our single-round search space does not support. The low ceiling there reflects the search space rather than the optimizer, and our agent still ranks best on MuSiQue.

\begin{figure}[t]
\centering
\includegraphics[width=\columnwidth]{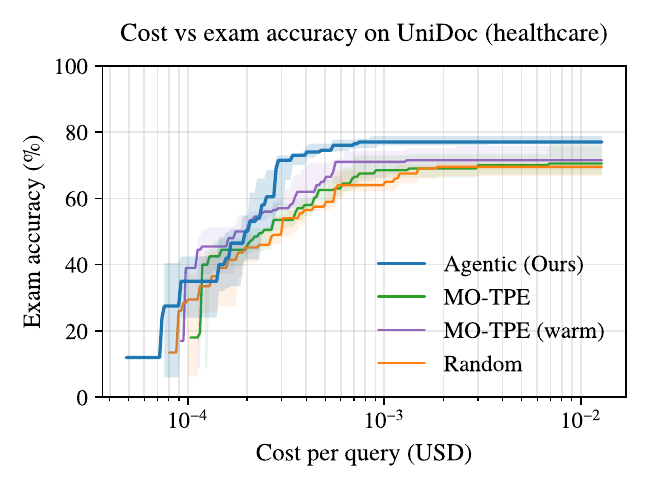}
\caption{Cost-quality results on the UniDoc healthcare corpus, aggregated over 10 seeds. Mean API cost per query on a log scale against accuracy on the self-generated exam, with each method drawn as the point-wise median across seeds of its cost-accuracy attainment (line) and the interquartile (25th to 75th percentile) spread across seeds (band). We compare our cost-aware optimizer (Agentic) against random search, cold MO-TPE, and warm-started MO-TPE.}
\label{fig:pareto}
\end{figure}

\subsection{Pareto experiment}
\label{sec:results:cost}

Figure~\ref{fig:pareto} traces each method's median cost-accuracy curve across the 10 seeds on the healthcare corpus. The agent's median curve reaches 77\% exam accuracy at \$0.000741 per query. The strongest baseline, warm-started MO-TPE, peaks at 71.5\% and pays \$0.001284, so the agent's best point is more accurate at 57.7\% of the cost. It matches that 71.5\% at \$0.000288, 22.4\% of what the baseline pays. It leads every baseline above \$0.000236 per query. Warm MO-TPE leads only in a band from \$0.000097 to \$0.000236, where neither curve reaches 60\% accuracy.

These median-curve gaps could still come from a few lucky seeds, so we test each claim across all 10 seeds, using two-sided Mann-Whitney U tests \citep{mann1947whitney} on per-seed quantities, Holm-corrected per family of three baseline comparisons \citep{holm1979simple}. The higher ceiling holds: the agent's per-seed peak exam accuracies span 74\% to 82\% and rank systematically above each baseline's (Holm-corrected $p = 0.017$ against warm MO-TPE, $0.025$ against cold MO-TPE, and $0.001$ against random search). The low-cost match holds as well: all ten agent seeds reach warm MO-TPE's median peak of 71.5\% within their 30 trials, against five of ten warm MO-TPE seeds, four of ten cold MO-TPE seeds, and two of ten random-search seeds. When seeds are ranked by the lowest per-query cost at which they reach this accuracy, with seeds that never reach it counted as the most expensive, the agent leads every baseline (Holm-corrected $p \leq 0.001$).

\section{Discussion}
\label{sec:discussion}

The two experiments show the optimizer from two angles. In the Accuracy experiment, which isolates the search from cost, the agent is sample-efficient: its best configuration within the first 10 trials already matches or beats every statistical baseline's full 30-trial held-out judge accuracy, at 0.817 against 0.786 on HotpotQA, 0.773 against 0.731 on MultiHop-RAG, and a tie on MuSiQue at 0.303 against 0.301, and the best-so-far trace (Figure~\ref{fig:best-so-far}) puts it near the top of every panel within the first few trials. The remaining 20 trials refine this region, and the full budget is what lets the agent match the stronger no-KB/diag ablation on MultiHop-RAG and lead on exact match and F1 on MuSiQue. The sample-efficiency reading holds even against warm-started MO-TPE, which begins with a free 30-trial same-corpus prior and so draws on information from 60 evaluations, yet is matched or beaten by the agent's first 10 counted trials.

The Pareto experiment is the setting we built the optimizer for, and it is a practical deployment scenario. Deploying a RAG pipeline is rarely a search for the most accurate configuration at any price. A practitioner usually wants the cheapest pipeline that clears an accuracy bar, because inference cost is paid on every query once the system is live. Our cost-aware experiment puts the agent in exactly this position, running it end to end on the real-world healthcare corpus. On this corpus the agent's median trade-off both undercuts the baselines on cost and beats them on accuracy: it matches the strongest baseline's median accuracy at about 22\% of the cost and extends to a higher ceiling that still costs less per query than the baselines' own best. It manages this by pricing and ranking candidates against its knowledge base before running them, so it moves directly to small but capable generators instead of spending trials to learn which cheap model suffices. Instead of a single configuration it returns the whole accuracy-cost trade-off and leaves the operating point to the user.

\section{Conclusion}
\label{sec:conclusion}

We introduce Agentic AutoRAG, a reasoning-driven optimizer for RAG pipelines. Instead of reducing each trial to an opaque score, it scores every configuration on a frozen corpus-derived exam, and after each trial a Diagnoser attributes each failed question to retrieval or generation while a Proposer, grounded in a knowledge base of public model rankings and pricing, selects the next configuration. On three multi-hop QA benchmarks it reaches higher LLM-judge accuracy than every baseline we compare and finds most of that accuracy within its first 10 trials. In its cost-aware mode on a healthcare corpus it reaches a median exam accuracy of 77\%, above the strongest baseline's 71.5\%, at about 58\% of that baseline's cost per query, and it matches that 71.5\% at about 22\% of the cost.

\section*{Limitations}

\paragraph{Fixed parser.}
Parsing is outside the search space. Every trial parses the corpus once with Docling, caches the result, and tunes chunking and downstream stages on top of it. Reparsing on every trial would dominate per-trial wall-clock time.

\paragraph{Knowledge-base inputs.}
The Proposer's KB encodes LLM rankings from Artificial Analysis and per-token pricing from LiteLLM, neither of which is peer-reviewed. The KB is also a point-in-time snapshot. Vendors re-rank, deprecate models, and update prices over time, so the Proposer's grounding is accurate at snapshot time and degrades with the gap to the run.

\paragraph{Agent nondeterminism.}
Inference temperature is configurable, but the agent models limit reproducibility. The Composer, Diagnoser, and Proposer default to a frontier reasoning model, several reasoning models require a temperature of 1.0 to function, and hosted reasoning-model APIs do not guarantee bit-exact outputs.

\paragraph{Judge dependence.}
Accuracy in both experiments is scored by an LLM judge, and we run no human-agreement study, so any systematic bias in the judge carries into the numbers we report.

\section*{Future Work}

\paragraph{Richer cross-document bridges.}
Our exam pipeline ranks bridges by lexical TF-IDF similarity, which connects only chunks that share rare surface terms. An entity-relation graph over the corpus, as in graph-based RAG \citep{edge2024graphrag}, would also surface bridges that share an entity but not its wording.

\paragraph{Expanding the search space.}
The diagnose-then-propose loop is agnostic to what the search space contains, so it can tune richer pipeline families than the modular flow we search here. Supporting whole architectures, such as agentic RAG or graph-based RAG, would let it search across retrieval paradigms rather than within one.

\paragraph{More objectives.}
The optimizer weighs only accuracy and cost today, and its knowledge base already records throughput and memory, so latency or memory footprint are natural additions.

\paragraph{Warm starts across runs.}
Each run begins cold on a new corpus. Carrying diagnoses and configuration priors across runs would let it start warm, accumulating an agent memory the optimizer retrieves from to generalize what it learns about RAG configuration from one corpus to the next.

\section*{Acknowledgments}

This work was carried out at the ETH Agentic Systems Lab. We thank David Schaurecker for providing API credits to support the LLM usage.

\bibliography{references}

\appendix

\section{Question taxonomy and exam samples}
\label{sec:appendix}

This appendix complements the exam-generation pipeline in Section~\ref{sec:method:examiner} by listing the seven question types the Composer draws from and showing one representative item per type. The examples come from the 100-question exam the optimizer generated for the UniDoc healthcare corpus, the one experiment where the generated exam is the optimization target. The Accuracy experiment instead optimizes against each benchmark's own held-in questions and generates no exam, so this appendix documents the pipeline where it runs.

\subsection{Question types}
\label{sec:appendix:taxonomy}

The Composer assigns one type per question from a fixed taxonomy of seven question types defined a priori. The taxonomy extends the bridge and comparison split of HotpotQA \citep{yang2018hotpotqa} with numeric and inference variants common to multi-hop benchmarks such as MuSiQue \citep{trivedi2022musique}.

\begin{description}
  \item[Extraction.] A factoid lookup answerable from a single span.
  \item[Definitional.] A short definition or description the chunk states explicitly.
  \item[Numeric (single-span).] Computes a value not stated verbatim by combining at least two numeric literals inside one span (sum, difference, range, median, count).
  \item[Inference.] Composes facts from at least two distinct spans into an answer that is not a contiguous substring (causal chain, implicit-referent resolution, qualitative direction from quantitative facts).
  \item[Bridge.] References an entity in one chunk via an indirect descriptor that another chunk uniquely identifies, then asks for an attribute of that entity.
  \item[Comparison.] Reads a comparable value from each cited chunk and returns the relation (which is larger, by how much).
  \item[Numeric (multi-span).] Computes across cited spans: difference, sum, ratio, duration.
\end{description}

Both numeric types also carry an arithmetic formula. The Composer must supply it alongside the answer, and the validator rejects the question when the formula does not evaluate to the canonical answer. This guards against arithmetic mistakes by the Composer LLM.

\paragraph{Source spans.}
Each source span is the minimal excerpt the answer strictly depends on, usually a single sentence. A span carries none of the surrounding context, so the example spans below can be hard to read on their own. Each records only the exact text a pipeline must retrieve, not the context that explains the answer. Retrieval counts as successful for a question only when the retrieved chunks contain all of its spans, and the per-question diagnosis uses this test to attribute a failure to weak retrieval rather than weak generation.

\paragraph{Distribution on UniDoc.}
Table~\ref{tab:unidoc-type-counts} reports the type distribution of the 100 questions that survive validation and probe-based selection on UniDoc. Extraction is well represented (16): the documents state many specific facts buried in long, multi-section text that are not trivially retrievable. Definitional questions do not survive selection, and only one single-span numeric question does, because most numeric reasoning here combines values that sit in different spans. The exam also stays within documents: 98 questions keep all their spans within one document, and only 2 cross into a second.

\begin{table}[t]
  \centering
  \small
  \begin{tabular}{lr}
    \toprule
    Question type & Count \\
    \midrule
    Extraction & 16 \\
    Definitional & 0 \\
    Numeric (single-span) & 1 \\
    Inference & 25 \\
    Bridge & 19 \\
    Comparison & 25 \\
    Numeric (multi-span) & 14 \\
    \midrule
    Total & 100 \\
    \bottomrule
  \end{tabular}
  \caption{Question-type distribution of the 100-question UniDoc exam.}
  \label{tab:unidoc-type-counts}
\end{table}

\subsection{Example questions}
\label{sec:appendix:examples}

One representative item per surviving type, drawn verbatim from the UniDoc exam. Source spans are reproduced as written.

\newtcolorbox{examitem}[1]{%
  enhanced, colback=gray!4, colframe=gray!40,
  colbacktitle=gray!4, coltitle=black,
  fonttitle=\bfseries\small, fontupper=\small,
  boxrule=0.6pt, titlerule=0.4pt,
  breakable, before skip=4pt, after skip=4pt,
  title={#1}%
}

\begin{examitem}{Extraction}
\textbf{Question.} What fraction of European American haplotypes in the registry are represented by only one individual?\\[2pt]
\textbf{Answer.} 44.65\% (variant: ``44.65 percent'').\\[2pt]
\textbf{Source span.} ``Application of our estimator to 6.59 million donors in the Be The Match Registry revealed that haplotypes follow a heavy tail distribution across all ethnicities: for example, 44.65\% of the European American haplotypes are represented by only 1 individual.''
\end{examitem}

\begin{examitem}{Numeric (single-span)}
\textbf{Question.} What is the average CK-MB value across both cattle cases with tropical theileriosis?\\[2pt]
\textbf{Answer.} 362.5 U/L.\\[2pt]
\textbf{Formula.} $(344 + 381) / 2 = 362.5$.\\[2pt]
\textbf{Source span.} ``CK-MB, Table 2. Biochemical findings of two cases.Unit = (U/L). CK-MB, Case-1 = 344. CK-MB, Case-2 = 381. CK-MB, Referance Range* = 189-235 **.''
\end{examitem}

\begin{examitem}{Inference}
\textbf{Question.} Why does the commercial Indian antivenom show limited efficacy against the newly discovered krait species from Southern India?\\[2pt]
\textbf{Answer.} Antivenom is manufactured from B. caeruleus venom, not from the new species (variant: ``because it is made from B. caeruleus venom'').\\[2pt]
\textbf{Source spans.}\\
``Considering the near countrywide distribution of this clinically important snake, B. caeruleus venoms are used for the manufacture of commercial Indian polyvalent antivenoms.''\\[4pt]
``While both antivenoms recognised the venoms of B. caeruleus (titre: 1:500; PSVPL: 1.02 Optical Density (OD); Haffkine: 0.89 OD) and B. sindanus (titre: 1:500; PSVPL: 0.89 OD; Haffkine: 0.76 OD) relatively well, they exhibited limited to poor recognition potential against the venom of B. romulusi (titre: 1:500; PSVPL: 0.24 OD; Haffkine: 0.25 OD).''
\end{examitem}

\begin{examitem}{Bridge}
\textbf{Question.} For the treatment comparison that showed a statistically significant difference in adverse reactions ($p<0.001$), what was the male-to-female ratio in each arm?\\[2pt]
\textbf{Answer.} 59:7 (variant: ``59/7'').\\[2pt]
\textbf{Source spans.}\\
``Male/female ratio, Group I (Budesonide 400 mcg plus Formoterol 6 mcg) = 59/7. Male/female ratio, Group II (Tiotropium 18 mcg plus Formoterol 12 mcg) = 59/7.''\\[4pt]
``In group I, following adverse drug reactions were observed; sore throat, hoarseness of voice, oral ulcers, tremors, 2 patients showed acute exacerbation and 1 had pneumonitis. In group II, following adverse drug reactions were observed; dryness of mouth, palpitations, tremors and 2 patients developed urinary retention. After that we have done Fisher's exact test and we observed that there is statistically significant difference ($p<0.001$) between group I and group II.''
\end{examitem}

\begin{examitem}{Comparison}
\textbf{Question.} How did particle concentration change when antibodies were added without magnetic beads, compared to when uncoated beads were used?\\[2pt]
\textbf{Answer.} Increased by 105\% versus no change (variant: ``rose by 105\% versus unchanged'').\\[2pt]
\textbf{Source spans.}\\
``We applied an identical bead procedure, except that we abstained from coating the beads with anti-ApoB antibodies prior to the incubation with the diluted PFP sample. This resulted in no changes in particle concentration and size distribution curve.''\\[4pt]
``we conducted a parallel experiment with a two-hour incubation of diluted PFP with 10 $\mu$g of anti-ApoB antibodies that had not been attached to magnetic beads, which resulted in an increase in particle concentration by 105\% as well as a notable shift of the size distribution curve to the right''
\end{examitem}

\begin{examitem}{Numeric (multi-span)}
\textbf{Question.} What is the ratio of items to care-plan domains in the interRAI Mental Health assessment?\\[2pt]
\textbf{Answer.} 19.8 items per domain (variant: ``approximately 20 items per domain'').\\[2pt]
\textbf{Formula.} $396 / 20 = 19.8$.\\[2pt]
\textbf{Source spans.}\\
``The interRAI Mental Health (MH) assessment (157, 158) supports care plan development in 20 domains, and is used in in-patient mental health settings''\\[4pt]
``The assessment is comprised of 396 items in several domains (see Table 1).''
\end{examitem}

\section{Search space}
\label{sec:appendix:searchspace}

The search space is one of the inputs the user supplies to the optimizer, alongside the corpus. It is a single YAML file (Listing~\ref{lst:config}) that names the corpus, the models that drive the optimizer, the exam-generation settings, and the dimensions to search. The optimizer reads this file directly, so the search space is fully specified and auditable before any trial runs.

Listing~\ref{lst:config} shows the configuration for the Pareto experiment on UniDoc, where the optimizer generates and scores its own exam. The Accuracy experiment, on HotpotQA, MuSiQue, and MultiHop-RAG, shares the same \texttt{search\_space} and differs only in its exam source, cost mode, and parsing, each described below. Those runs supply each benchmark's own held-in questions as the validation exam through \texttt{examiner.custom\_exam\_path} and generate none, so the \texttt{examiner} block in the listing does not apply to them. They also run score-only rather than cost-aware, and they parse pre-extracted text with OCR and table-structure extraction off. UniDoc, built from real PDFs and page images, turns both on, sets \texttt{cost\_aware} to true, and weights the exam-generation neighborhoods toward same-document siblings (\texttt{0.8 / 0.2}) because its documents are long and multi-section.

\begin{lstlisting}[style=yamlpaper,caption={Project configuration for the Pareto experiment on UniDoc, the file the user supplies to the optimizer. The Accuracy experiment shares the same \texttt{search\_space} but supplies its own held-in exam and runs score-only.},label=lst:config]
# Project configuration: corpus, optimizer models, exam
# settings, and the search space. Provided by the user.

meta:
  project_name: unidoc-pareto
  corpus_path: ./corpus
  # free-text hint read by the optimizer
  corpus_description: |
    Real-world healthcare PDFs and scanned page images:
    insurance forms, claims, clinical notes. Dense,
    domain-specific prose with tables and mixed layouts.
  max_trials: 30   # search budget
  cost_aware: true   # Pareto run; the Accuracy experiment sets false

model_aliases:   # map names to your deployments
  azure/gpt-5-nano:   <your-deployment>
  azure/gpt-5-mini:   <your-deployment>
  azure/gpt-4.1-nano: <your-deployment>
  azure/gpt-4.1-mini: <your-deployment>

# LLMs that drive the optimizer and grade answers
agent:
  optimizer_model: kimi-k2.5      # diagnose + propose
  examiner_model:  minimax-m2.5   # compose exam
  judge_model:     DeepSeek-V3.2  # grade answers
  optimizer_reasoning_effort: medium
  examiner_reasoning_effort: medium

examiner:   # exam generation (skipped when an exam is supplied)
  exam_size: 100
  # seed this x exam_size neighborhoods; the composer
  # over-generates questions, then filter to exam_size
  initial_question_multiplier: 1.8
  probe_selection: true   # keep discriminative items
  neighborhood_min_chunks: 18   # context per call
  neighborhood_min_words: 5000
  # neighborhood mix, favoring same-document spans
  neighborhood_same_doc_weight: 0.8
  neighborhood_cross_doc_weight: 0.2

parsing:   # OCR + tables on for the UniDoc PDFs
  parser: docling
  ocr: true
  table_structure: true

search_space:   # dimensions searched
  chunking:   # how documents are split
    strategies: [recursive, fixed]
    chunk_token_size: [128, 256, 384, 512]
    chunk_token_overlap: [0, 32, 48, 64]

  embedding:   # 13 retrievers
    models:
      - IEITYuan/Yuan-embedding-2.0-en
      - Qwen/Qwen3-Embedding-0.6B
      - ibm-granite/granite-embedding-english-r2
      - ibm-granite/granite-embedding-small-english-r2
      - intfloat/e5-large-v2
      - intfloat/multilingual-e5-large-instruct
      - BAAI/bge-large-en-v1.5
      - BAAI/bge-small-en-v1.5
      - BAAI/bge-m3
      - Snowflake/snowflake-arctic-embed-l-v2.0
      - Snowflake/snowflake-arctic-embed-xs
      - mixedbread-ai/mxbai-embed-large-v1
      - sentence-transformers/all-MiniLM-L6-v2

  retrieval:
    index_types: [vector_only, hybrid_bm25_vector]
    top_k: { min: 3, max: 20 }   # retrieved chunks
    hybrid_alpha: { min: 0.0, max: 1.0 }   # lexical/dense
    bm25_vector_fusion: [alpha, rrf]
    long_context_reorder: [false, true]

  query_expansion:
    strategies: [none, hyde, multi_query, query_decompose]
    models: [azure/gpt-4o-mini]   # fixed expander

  reranker:
    models:
      - none
      - BAAI/bge-reranker-v2-m3
      - cross-encoder/ms-marco-MiniLM-L-6-v2
      - Alibaba-NLP/gte-reranker-modernbert-base
      - mixedbread-ai/mxbai-rerank-xsmall-v1
    top_n: { min: 3, max: 10 }   # reranked chunks

  generator:   # 35 answer LLMs
    models:
      - bedrock/us.amazon.nova-micro-v1:0
      - bedrock/us.amazon.nova-lite-v1:0
      - bedrock/us.amazon.nova-pro-v1:0
      - bedrock/google.gemma-3-4b-it
      - bedrock/google.gemma-3-12b-it
      - bedrock/google.gemma-3-27b-it
      - bedrock/us.meta.llama3-1-8b-instruct-v1:0
      - bedrock/us.meta.llama3-3-70b-instruct-v1:0
      - bedrock/us.meta.llama4-scout-17b-instruct-v1:0
      - bedrock/us.meta.llama4-maverick-17b-instruct-v1:0
      - bedrock/mistral.magistral-small-2509
      - bedrock/mistral.ministral-3-3b-instruct
      - bedrock/mistral.ministral-3-8b-instruct
      - bedrock/mistral.ministral-3-14b-instruct
      - bedrock/mistral.mistral-7b-instruct-v0:2
      - bedrock/mistral.mistral-large-3-675b-instruct
      - bedrock/mistral.mixtral-8x7b-instruct-v0:1
      - bedrock/nvidia.nemotron-nano-3-30b
      - bedrock/openai.gpt-oss-20b-1:0
      - bedrock/openai.gpt-oss-120b-1:0
      - bedrock/qwen.qwen3-32b-v1:0
      - bedrock/qwen.qwen3-coder-30b-a3b-v1:0
      - bedrock/zai.glm-4.7
      - azure/gpt-4o
      - azure/gpt-4o-mini
      - azure/gpt-4.1-mini
      - azure/gpt-4.1-nano
      - azure/o4-mini
      - azure/gpt-5-mini
      - azure/gpt-5-nano
      - azure/gpt-5.4-mini
      - azure/gpt-5.4-nano
      - vertex_ai/gemini-2.5-flash
      - vertex_ai/gemini-2.5-flash-lite
      - vertex_ai/gemini-3.1-flash-lite-preview
    # when true, trials may use reasoning or plain models,
    # passing reasoning params to LiteLLM where supported
    reasoning: true
    reasoning_effort: medium   # when reasoning is on

  temperature: { min: 1.0, max: 1.0 }
\end{lstlisting}

\section{Knowledge base}
\label{sec:appendix:kb}

Appendix~\ref{sec:appendix:searchspace} names the candidate models. This appendix shows the annotations the Proposer reasons over for each one. The knowledge base pairs every model in the search space with public benchmark scores, per-token pricing, and serving attributes, which lets the Proposer rank candidates and rule out quality-cost-dominated choices before it spends a trial on them. LLM benchmarks and throughput come from Artificial Analysis,\footnote{\url{https://artificialanalysis.ai/}} embedding and reranker rankings from the MTEB \citep{muennighoff2023mteb}, MMTEB \citep{enevoldsen2025mmteb}, and FollowIR \citep{weller2025followir} benchmarks, and per-token pricing from LiteLLM.\footnote{\url{https://github.com/BerriAI/litellm}} The optimizer renders the knowledge base filtered to the configured search space and to the run's reasoning-effort setting, so a model that supports reasoning can appear as two entries, one reasoning and one non-reasoning, with different scores. The 35 generator models of Appendix~\ref{sec:appendix:searchspace} therefore render as 49 scored entries.

Tables~\ref{tab:kb-llm}, \ref{tab:kb-embed}, and \ref{tab:kb-rerank} report the generator, embedding, and reranker annotations. The embedding and reranker tables are complete, whereas for generators we show twelve models spanning the intelligence range. The Proposer treats these scores as comparative signal, not ground truth.

The scores are a point-in-time snapshot, taken in May 2026 for the LLM benchmarks and pricing and from the MTEB(eng, v2) leaderboard for the embedders.

\begin{table*}[t]
  \centering
  \small
  \begin{tabular*}{\textwidth}{@{\extracolsep{\fill}}llrrrrrrc}
    \toprule
    Model & Creator & Intel. & GPQA & IFBench & In \$/M & Out \$/M & Tok/s & Reasoning \\
    \midrule
    glm-4.7          & Z AI    & 42.1 & 0.859 & 0.679 & 0.60 & 2.20  & 86  & Yes \\
    gpt-5-mini       & OpenAI  & 38.9 & 0.803 & 0.712 & 0.25 & 2.00  & 106 & Yes \\
    gpt-5.4-mini     & OpenAI  & 37.7 & 0.823 & 0.648 & 0.75 & 4.50  & 172 & Yes \\
    gpt-oss-120b     & OpenAI  & 33.3 & 0.782 & 0.690 & 0.15 & 0.60  & 289 & Yes \\
    gemini-2.5-flash & Google  & 27.0 & 0.790 & 0.503 & 0.30 & 2.50  & 225 & Yes \\
    gpt-5-nano       & OpenAI  & 25.9 & 0.670 & 0.659 & 0.05 & 0.40  & 151 & Yes \\
    gpt-4.1-mini     & OpenAI  & 22.9 & 0.664 & 0.383 & 0.40 & 1.60  & 80  & No  \\
    mistral-large-3  & Mistral & 22.8 & 0.680 & 0.362 & 0.50 & 1.50  & 58  & No  \\
    llama4-maverick  & Meta    & 18.4 & 0.671 & 0.430 & 0.24 & 0.97  & 122 & No  \\
    gpt-4o           & OpenAI  & 17.3 & 0.543 & 0.343 & 2.50 & 10.00 & 172 & No  \\
    nova-micro       & Amazon  & 10.3 & 0.358 & 0.294 & 0.04 & 0.14  & 318 & No  \\
    gemma-3-4b       & Google  & 6.3  & 0.291 & 0.283 & 0.04 & 0.08  & \textendash & No  \\
    \bottomrule
  \end{tabular*}
  \caption{Generator-LLM annotations in the knowledge base, a twelve-model slice across the intelligence range (the full set of 35 models, which render as 49 reasoning and non-reasoning entries, is named in Appendix~\ref{sec:appendix:searchspace}). Intel.\ is the Artificial Analysis intelligence index. GPQA \citep{rein2024gpqa} and IFBench \citep{pyatkin2025ifbench} are accuracies in $[0,1]$. Prices are USD per million input and output tokens. Tok/s is median output throughput. Reasoning marks which variant the listed scores describe. Model names abbreviate the LiteLLM identifiers in Appendix~\ref{sec:appendix:searchspace}, and a dash marks a value unpublished in the snapshot.}
  \label{tab:kb-llm}
\end{table*}

\begin{table*}[t]
  \centering
  \small
  \begin{tabular*}{\textwidth}{@{\extracolsep{\fill}}lrrrrrrr}
    \toprule
    Model & Dim. & Max tok. & Params (B) & Retrieval & STS & Rerank. & Mem. (MB) \\
    \midrule
    Yuan-embedding-2.0-en              & 1024 & 2048  & 0.596 & 0.707 & 0.849 & 0.533 & 2272 \\
    Qwen3-Embedding-0.6B               & 1024 & 32768 & 0.596 & 0.618 & 0.858 & 0.482 & 1136 \\
    snowflake-arctic-embed-l-v2.0      & 1024 & 8192  & 0.568 & 0.586 & 0.770 & 0.475 & 2166 \\
    granite-embedding-english-r2       & 768  & 8192  & 0.149 & 0.564 & 0.781 & 0.491 & 284  \\
    mxbai-embed-large-v1               & 1024 & 512   & 0.335 & 0.556 & 0.772 & 0.480 & 639  \\
    bge-large-en-v1.5                  & 1024 & 512   & 0.335 & 0.554 & 0.759 & 0.483 & 1242 \\
    granite-embedding-small-english-r2 & 384  & 8192  & 0.048 & 0.539 & 0.766 & 0.476 & 91   \\
    bge-small-en-v1.5                  & 512  & 512   & 0.033 & 0.539 & 0.739 & 0.469 & 127  \\
    multilingual-e5-large-instruct     & 1024 & 514   & 0.560 & 0.535 & 0.843 & 0.487 & 1068 \\
    snowflake-arctic-embed-xs          & 384  & 512   & 0.023 & 0.526 & 0.728 & 0.453 & 86   \\
    bge-m3                             & 1024 & 8194  & 0.568 & 0.510 & 0.799 & \textendash & 2167 \\
    e5-large-v2                        & 1024 & 514   & 0.335 & 0.493 & 0.749 & 0.457 & 1278 \\
    all-MiniLM-L6-v2                   & 384  & 256   & 0.023 & 0.430 & 0.704 & 0.471 & 87   \\
    \bottomrule
  \end{tabular*}
  \caption{The 13 embedding models in the search space with their MTEB(eng, v2) \citep{muennighoff2023mteb} scores. Retrieval, STS, and reranking (Rerank.)\ are the corresponding MTEB task scores in $[0,1]$. Dim.\ is the embedding dimension, Max tok.\ the maximum input length, Params (B) the parameter count in billions, and Mem.\ the memory footprint. The Proposer ranks retrievers by the Retrieval score. Names drop the organization prefix listed in Appendix~\ref{sec:appendix:searchspace}, and a dash marks a value unpublished in the snapshot.}
  \label{tab:kb-embed}
\end{table*}

\begin{table*}[t]
  \centering
  \small
  \begin{tabular*}{\textwidth}{@{\extracolsep{\fill}}lrrrr}
    \toprule
    Model & Params (B) & MTEB-R & MMTEB-R & FollowIR \\
    \midrule
    bge-reranker-v2-m3            & 0.6   & 0.570 & 0.584 & $-0.01$ \\
    ms-marco-MiniLM-L-6-v2        & 0.022 & \textendash & \textendash & \textendash \\
    gte-reranker-modernbert-base  & \textendash & \textendash & \textendash & \textendash \\
    mxbai-rerank-xsmall-v1        & \textendash & \textendash & \textendash & \textendash \\
    \bottomrule
  \end{tabular*}
  \caption{The four reranker models in the search space (a fifth option, \texttt{none}, disables reranking). MTEB-R and MMTEB-R are the MTEB \citep{muennighoff2023mteb} and MMTEB \citep{enevoldsen2025mmteb} reranking-task scores in $[0,1]$, and FollowIR \citep{weller2025followir} is the instruction-following retrieval score. Coverage is sparse, with published scores for only one of the four rerankers in the snapshot, so the Proposer has little benchmark signal here and relies on rerank depth and downstream accuracy. Names abbreviate the identifiers in Appendix~\ref{sec:appendix:searchspace}, and a dash marks a value unpublished in the snapshot.}
  \label{tab:kb-rerank}
\end{table*}

\clearpage
\twocolumn[%
  \noindent{\includegraphics[width=\textwidth]{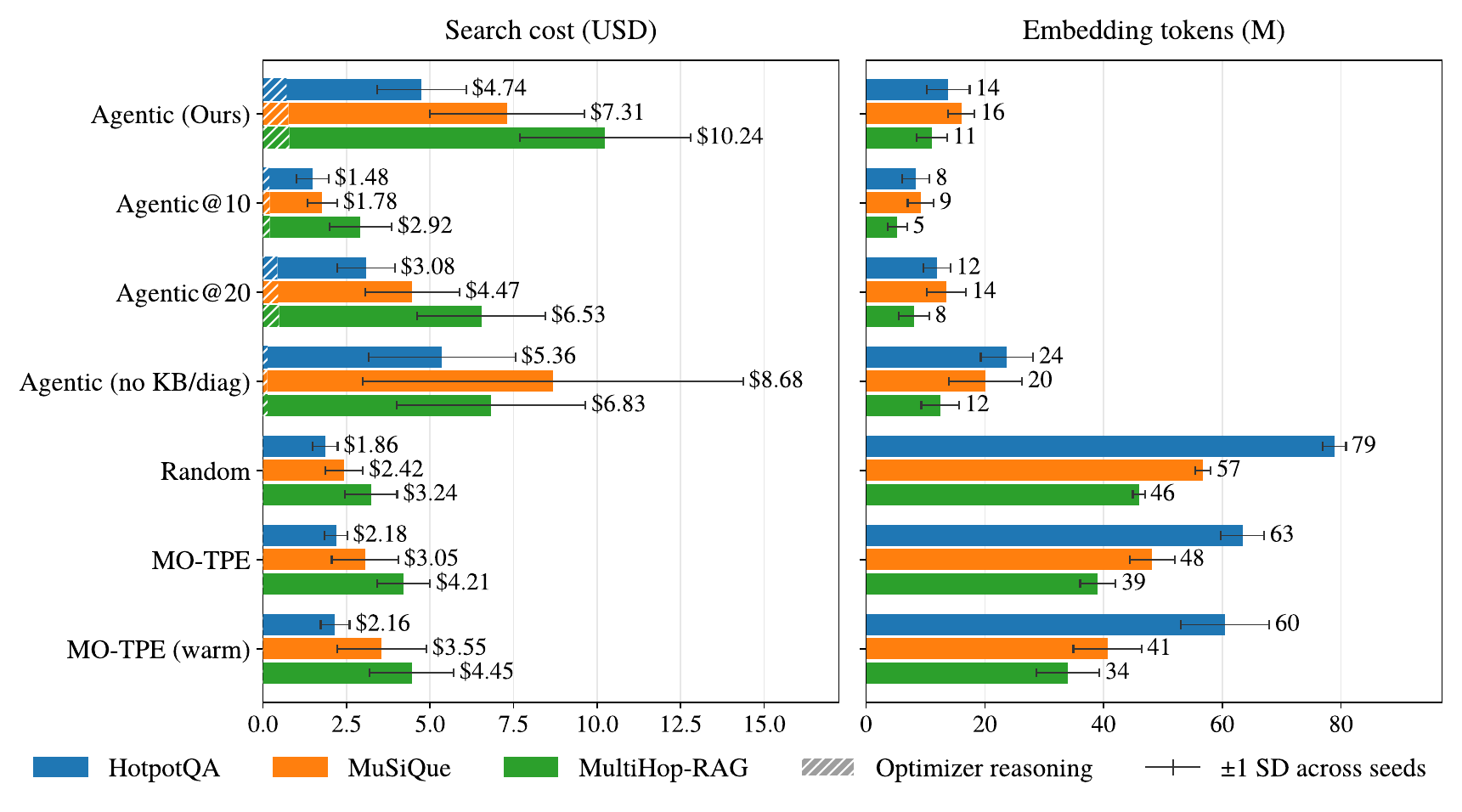}}\par
  \captionof{figure}{Search cost and embedding compute on the three datasets of the Accuracy experiment. Bars show the mean over the 10 search seeds and whiskers span $\pm 1$ standard deviation across seeds. In the search-cost panel each bar splits into optimizer reasoning (the agent's Proposer and Diagnoser calls) and trial evaluation (the RAG generation and judge scoring at each trial). Random search and MO-TPE propose configurations without a language model, so their cost is entirely trial evaluation. The embedding panel counts embedding-input tokens in millions, one charge per distinct embedder and chunk-size configuration rather than per trial.}%
  \label{fig:cost-emb}%
  \vspace{\baselineskip}
]
\section{Search cost and embedding compute}
\label{sec:appendix:cost}

Figure~\ref{fig:cost-emb} reports two costs of running each optimizer on the three datasets of the Accuracy experiment, the dollar cost of the search and the embedding compute it consumes. We count the latter as embedding-input tokens, charged once per distinct embedder and chunk-size configuration, since each index is built once and reused across a run's trials. Bars are the mean over the 10 search seeds and whiskers span $\pm 1$ standard deviation across seeds. The methods differ most on embedding tokens used. The agent's search consumes 11 to 16M embedding tokens per dataset, against 46 to 79M for random search and 34 to 63M for MO-TPE, three to five times fewer. Guided by the knowledge base, it settles on a few strong embedders early and reuses their cached indexes, while random search and MO-TPE sample embedders and chunk sizes across the whole space and pay for a fresh index build on each new combination.

In dollars the methods are closer, and the full 30-trial agent costs more than the statistical baselines. The gap is trial evaluation, not reasoning. Its Proposer and Diagnoser calls add only \$0.71 to \$0.79 per run, while random search and MO-TPE choose configurations without a language model and spend nothing on reasoning. What raises the agent's bill is the pipelines it evaluates. These are score-only runs, so cost is not an objective and nothing steers the agent toward cheaper generators. It knows from the knowledge base and the language model's own priors which generators are strong, and strong generators tend to cost more per token, so its trials favor more capable, more expensive configurations. However, reaching their accuracy does not take the full budget. Its 10-trial checkpoint recovers most of the agent's final held-out accuracy (Table~\ref{tab:holdout}) at \$1.48 to \$2.92 per run, comparable to random search and MO-TPE at \$1.86 to \$4.45.

\end{document}